\documentclass[11pt,a4paper]{article}
\usepackage[hyperref]{naaclhlt2019}
\usepackage{times}
\usepackage{latexsym}
\usepackage{graphicx}
\usepackage{amsmath}
\usepackage{ragged2e} 
\usepackage{makecell, multirow, tabularx}
\newcolumntype{L}{>{\RaggedRight}X}
\graphicspath{ {images/} }
\usepackage{url}

\aclfinalcopy 

\title{Memory-Augmented Recurrent Networks for Dialogue Coherence}

\author{David Donahue, Yuanliang Meng, Anna Rumshisky \\
        {\tt\small david\_donahue@student.uml.edu, \{ymeng,arum\}@cs.uml.edu} \\
        Department of Computer Science \\
        University of Massachusetts Lowell \\
        Lowell, MA 01854
}

\begin{document}
\maketitle
\begin{abstract}

Recent dialogue approaches operate by reading each word in a conversation history, and aggregating accrued dialogue information into a single state. This fixed-size vector is not expandable and must maintain a consistent format over time. Other recent approaches exploit an attention mechanism to extract useful information from past conversational utterances, but this introduces an increased computational complexity. In this work, we explore the use of the Neural Turing Machine (NTM) to provide a more permanent and flexible storage mechanism for maintaining dialogue coherence. Specifically, we introduce two separate dialogue architectures based on this NTM design. The first design features a sequence-to-sequence architecture with two separate NTM modules, one for each participant in the conversation. The second memory architecture incorporates a single NTM module, which stores parallel context information for both speakers. This second design also replaces the sequence-to-sequence architecture with a neural language model, to allow for longer context of the NTM and greater understanding of the dialogue history. We report perplexity performance for both models, and compare them to existing baselines.

\end{abstract}

\section{Introduction}

Recently, chit-chat dialogue models have achieved improved performance in modelling a variety of conversational domains, including movie subtitles, Twitter chats and help forums \cite{vinyals2015neural, serban2016building, serban2017hierarchical, kingma2013auto}. These neural systems were used to model conversational dialogue via training on large chit-chat datasets such as the OpenSubtitles corpus, which contains generic dialogue conversations from movies \cite{lison2016opensubtitles2016}. The datasets used do not have an explicit dialogue state to be modelled \cite{ren2018towards}, but rather require the agent to learn the nuances of natural language in the context of casual peer-to-peer interaction.

Many recent chit-chat systems \cite{serban2017hierarchical, kingma2013auto} attempt to introduce increased diversity into model responses. However, dialogue systems have also been known to suffer from a lack of coherence \cite{vinyals2015neural}. Given an input message history, systems often have difficulty tracking important information such as professions and names \cite{vinyals2015neural}. It would be of benefit to create a system which extracts relevant features from the input that indicate which responses would be most appropriate, and conditions on this stored information to select the appropriate response. 



\begin{figure*}[ht]
\begin{center}
\includegraphics[scale=.3]{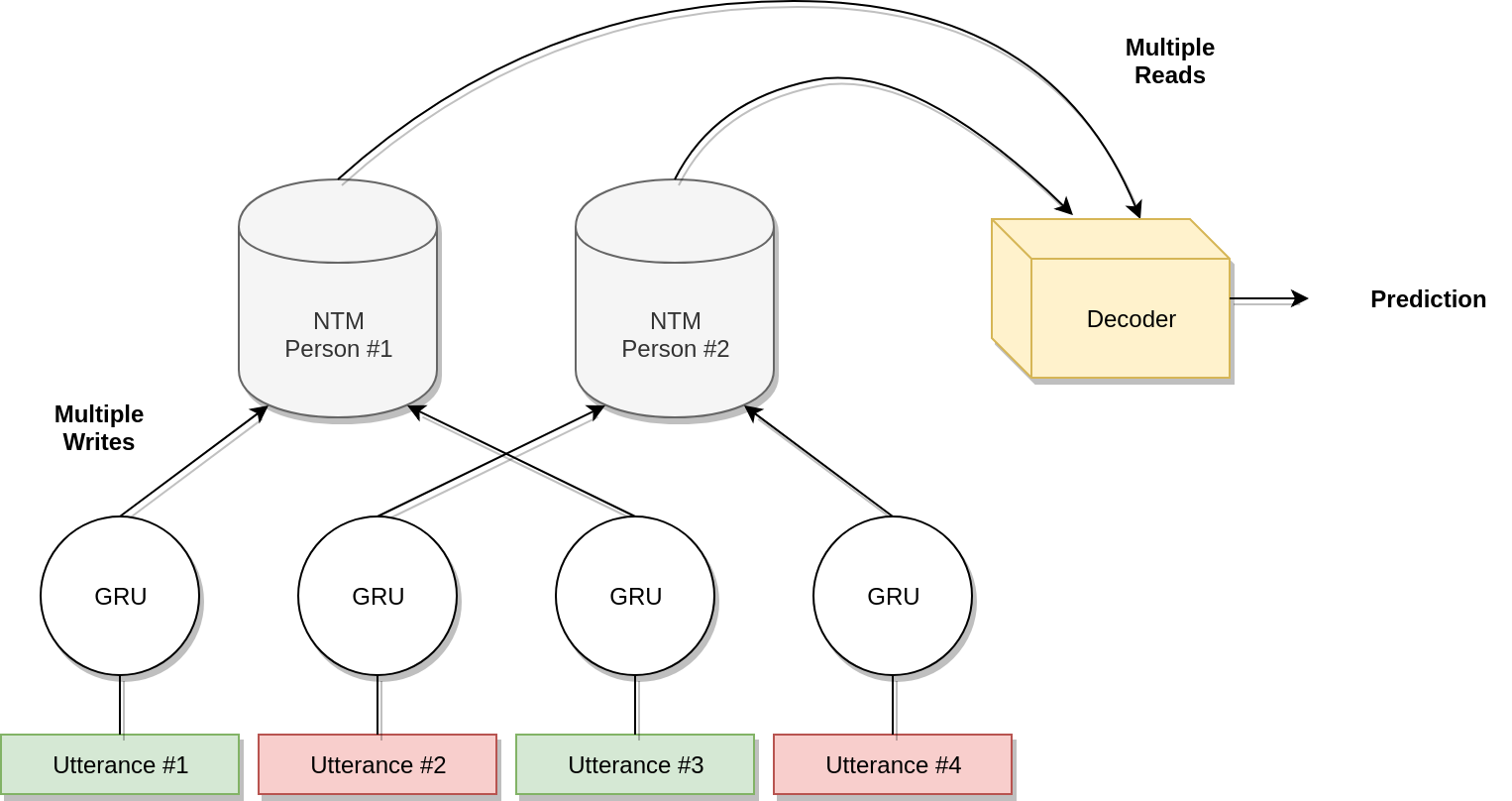}
\end{center}
\caption{Memory-augmented dialogue architecture with dual NTMs (D-NTMS). GRU encoders read each utterance in the conversation in segments. After reading each segment, a write is made to the corresponding Neural Turing Machine memory module (NTM). Two NTMs are designated, one for each speaker in the conversation (two speakers total). The resulting NTMs are read from and their predictions are used to output the final response prediction.}
\label{fig:memory}
\end{figure*}


A major problem with existing recurrent neural network (RNN) architectures is that these systems aggregate all input tokens into a state vector, which is passed to a decoder for generation of the final response, or in the case of a neural probabilistic language model \cite{bengio2001neural}, the state at each time step is used to predict the next token in the sequence. Ideally the size of the state should expand with the number of input tokens and should not lose important information about the input. However, RNN states are typically fixed sized, and for any chosen state size, there exists an input sequence length for which the RNN would not be able to store all relevant details for a final response. In addition, the RNN state undergoes constant transformation at each computational step. This makes it difficult to maintain a persistent storage of information that remains constant over many time steps.

The introduction of attention mechanisms \cite{bahdanau2014neural} has sparked a change in the current design of RNN architectures. Instead of relying fully on a fixed-sized state vector, an attention mechanism allows each decoder word prediction step to extract relevant information from past states through a key-value query mechanism. However, this mechanism connects every input token with all preceeding ones via a computational step, increasing the complexity of the calculation to $O(N^2)$ for an input sequence size N. In the ideal case, the mapping of input conversation history to output response would have a computational complexity of $O(N)$. For this reason, it is desirable to have an information retrieval system that is both scale-able, but not proportional to input length. 

We study the impact of accessible memory on response coherence by constructing a memory-augmented dialogue system. The motivation is that it would be beneficial to store details of the conversational history in a more permanent memory structure, instead of being captured inside a fixed-sized RNN hidden state. Our proposed system is able to both read and write to a persistent memory module after reading each input utterance. As such, it has access to a stable representation of the input message history when formulating a final response. We explore two distinct memory architectures with different properties, and compare their differences and benefits. We evaluate our proposed memory systems using perplexity evaluation, and compare them to competitive baselines.

\section{Recent Work}

\newcite{vinyals2015neural} train a sequence-to-sequence LSTM-based dialogue model on messages from an IT help-desk chat service, as well as the OpenSubtitles corpus, which contains subtitles from popular movies. This model was able to answer philosophical questions and performed well with common sense reasoning. Similarly, \newcite{serban2016building} train a hierarchical LSTM architecture (HRED) on the MovieTriples dataset, which contains examples of the form (utterance \#1, utterance \#2, utterance \#3). However, this dataset is small and does not have conversations of larger length. They show that using a context recurrent neural network (RNN) to read representations at the utterance-level allows for a more top-down perspective on the dialogue history. Finally, \newcite{serban2017hierarchical} build a dialogue system which injects diversity into output responses (VHRED) through the use of a latent variable for variational inference \cite{kingma2013auto}. They argue that the injection of information from the latent variables during inference increases response coherence without degrading response quality. They train the full system on the Twitter Dialogue corpus, which contains generic multi-turn conversations from public Twitter accounts. They also train on the Ubuntu Dialogue Corpus, a collection of multi-turn vocabulary-rich conversations extracted from Ubuntu chat logs. \newcite{du2018variational} adapt from the VHRED architecture by increasing the influence of the latent variables on the output utterance. In this work, a backwards RNN carries information from future timesteps to present ones, such that a backward state contains a summary of all future utterances the model is required to generate. The authors constrain this backward state at each time step to be a latent variable, and minimize the KL loss to restrict information flow. At inference, all backward state latent variables are sampled from and decoded to the output response. The authors interpret the sampling of the latent variables as a "plan" of what to generate next.

\newcite{bowman2015generating} observe that latent variables can sometimes degrade, where the system chooses not to store information in the variable and does not condition on it when producing the output. \newcite{bowman2015generating} introduce a process called KL-annealing which slowly increases the KL divergence loss component over the course of training. However, \cite{park2018hierarchical} claim that KL annealing is not enough, and introduce utterance dropout to force the model to rely on information stored in the latent variable during response generation. They apply this system to conversational modelling.

Other attempts to increase diversity focus on selecting diverse responses after the model is trained. \newcite{li2015diversity} introduce a modification of beam search. Beam search attempts to find the highest probability response to a given input by producing a tree of possible responses and "pruning" branches that have the lowest probability. The top K highest probability responses are returned, of which the highest is selected as the output response. \newcite{li2015diversity} observe that beam search tends to select certain families of responses that temporarily have higher probability. To combat this, a discount factor of probabilities is added to responses that come from the same parent response candidate. This encourages selecting responses that are different from one another when searching for the highest probability target.

While coherence and diversity remain the primary focus of model dialogue architectures, many have tried to incorporate additional capabilities. \newcite{zhou2017mojitalk} introduce emotion into generated utterances by creating a large-scale fine-grained emotion dialogue dataset that uses tagged emojis to classify utterance sentiment. Then they train a conditional variational autoencoder (CVAE) to generate responses given an input emotion. Along this line of research, \newcite{li2016persona} use Reddit users as a source of persona, and learn individual persona embeddings per user. The system then conditions on these embeddings to generate a response while maintaining coherence specific to the given user. \newcite{pandey2018exemplar} expand the context of an existing dialogue model by extracting input responses from the training set that are most similar to the current input. These "exemplar" responses are then conditioned on to use as reference for final response generation. In another attempt to add context, \newcite{young2018augmenting} utilize a relational database to extract specific entity relations that are relevant for the current input. These relations provide more context for the dialogue model and allows it to respond to the user with information it did not observe in the training set.

Ideally, NLP models should have the ability to use and update information processed in the past. For dialogue generation, this ability is particularly important, because dialogue involves exchange of information in discourse, and all responses depend on what has been mentioned in the past. 
RNNs introduce "memory" by adding an output of one time step to their input in a future time step. Theoretically, properly trained RNNs are Turing-complete, but in reality vanilla RNNs often do not perform well due to the gradient vanishing problem. Gated RNNs such as LSTM and GRU introduces cell state, which can be understood as memory controlled by trainable logic gates. Gated RNNs do not suffer from the vanishing gradient problem as much, and indeed outperform vanilla RNNs in various NLP tasks. This is likely because the vanilla RNN state vector undergoes a linear transformation at each step, which can be difficult to control. In contrast, gated RNNs typically both control the flow of information, and ensure only elemnt-wise operations occur on the state, which allow gradients to pass more easily. However, they too fail in some basic memorization tasks such as copying and associative recall. A major issue is when the cell state gets updated, previous memories are forever erased. As a result, Gated RNNs can not model long-term dependencies well.

In recent years, there have been proposals to use memory neural networks to capture long-term information. A memory module is defined as an external component of the neural network system, and it is theoretically unlimited in capacity. \newcite{weston2014memory} propose a sequence prediction method using a memory with content-based addressing. In their implementation for the bAbI task~\cite{WestonBCM15} for example, their model encodes and sequentially saves words from text in memory slots. When a question about the text is asked, the model uses content-based addressing to retrieve memories relevant to the question, in order to generate answers. They use the k-best memory slots, where k is a relative small number (1 or 2 in their paper). \newcite{sukhbaatar2015end} propose an end-to-end neural network model, which uses content-based addressing to access multiple memory layers. This model has been implemented in a relatively simple goal-oriented dialogue system (restaurant booking) and has decent performance~\cite{DBLP:journals/corr/BordesW16}.

\newcite{DBLP:journals/corr/GravesWD14} further develop the addressing mechanism and make old memory slots dynamically update-able. The model read heads access information from all the memory slots at once using soft addressing. The write heads, on the other hand, have the ability to modify memory slots. The content-based addressing serves to locate relevant information from memory, while another location-based addressing is also used, to achieve slot shifting, interpolation of address from the previous step, and so on. As a result, the memory management is much more complex than the previously proposed memory neural networks. This system is known as the Neural Turing Machine (NTM).

Other NTM variants have also been proposed recently. \newcite{DBLP:journals/corr/ZhangYZ15} propose structured memory architectures for NTMs, and argue they could alleviate overfitting and increase predictive accuracy. \newcite{DBLP:journals/nature/GravesWRHDGCGRA16} propose a memory access mechanism on top of NTM, which they call the Differentiable Neural Computer (DNC). DNC can store the transitions between memory locations it accesses, and thus can model some structured data.  \newcite{DBLP:journals/corr/GulcehreCCB16} proposed a Dynamic Neural Turing Machine (D-NTM) model, which allows more addressing mechanisms, such as multi-step addressing. \newcite{DBLP:journals/corr/GulcehreCB17} further simplified the algorithm, so a single trainable matrix is used to get locations for read and write. Both models separate the address section from the content section of memory. 

The Global Context Layer~\cite{P18-1049} independently proposes the idea of address-content separation, noting that the content-based addressing in the canonical NTM model is difficult to train. A crucial difference between GCL and these models is that they use input “content” to compute keys. In GCL, the addressing mechanism fully depends on the entity representations, which are provided by the context encoding layers and not computed by the GCL controller. Addressing then involves matching the input entities and the entities in memory. Such an approach is desirable for tasks like event temporal relation classification, entity co-reference and so on. GCL also simplified the location-based addressing proposed in NTM. For example, there is no interpolation between current addressing and previous addressing. 

Other than NTM-based approaches, there are recent models that use an attention mechanism over either input or external memory. For instance, the Pointer Networks~\cite{NIPS2015_5866} uses attention over input timesteps. However, it has no power to rewrite information for later use, since they have no “memory” except for the RNN states. The Dynamic Memory Networks~\cite{pmlr-v48-kumar16} have an “episodic memory” module which can be updated at each timestep. However, the memory is a vector (“episode”) without internal structure, and the attention mechanism only works on inputs, just as in Pointer Networks. The GCL model and other NTM-based models have a memory with multiple slots, and the addressing function dictates writing and reading to/from certain slots in the memory

\begin{figure*}[ht]
\begin{center}
\includegraphics[scale=.3]{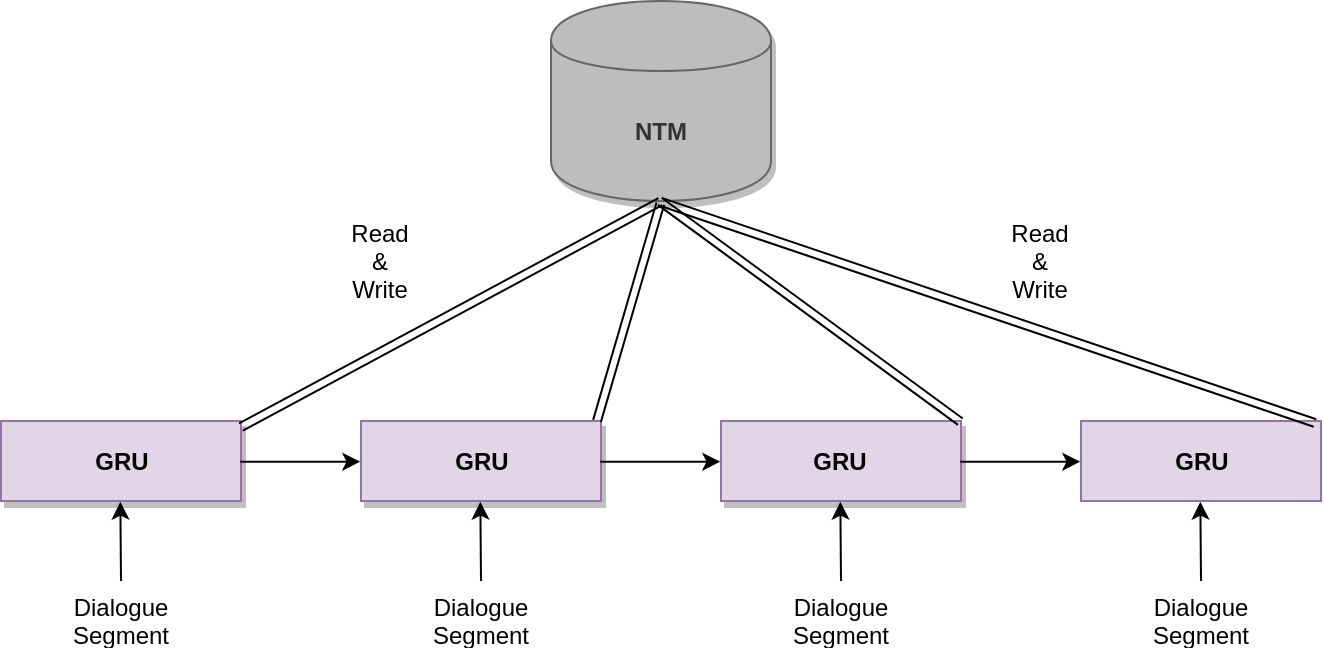}
\end{center}
\caption{Proposed single-NTM language model dialogue system (NTM-LM). The input dialogue history is broken into segments and each is processed by a GRU language model in sequence. At the end of each segment, the GRU state is used to read from and write to the persistent Neural Turing Machine (NTM).}
\label{fig:model}
\end{figure*}

\begin{table*}[ht]
\centering
\begin{tabular}{ |r|r| } 
 \hline
 Architecture & Perplexity \\

 \hline
 Seq2Seq & 75.44\\
 D-NTMS & 74.07\\
 HRED & 73.33\\
 LM & 69.36\\
 NTM-LM & \bf{68.50}\\

 \hline
\end{tabular}
\caption{\footnotesize Word-level perplexity evaluation on proposed model and two selected baselines.}
\label{table:eval}
\end{table*}

\begin{figure*}[ht]
\begin{center}
\includegraphics[scale=.3]{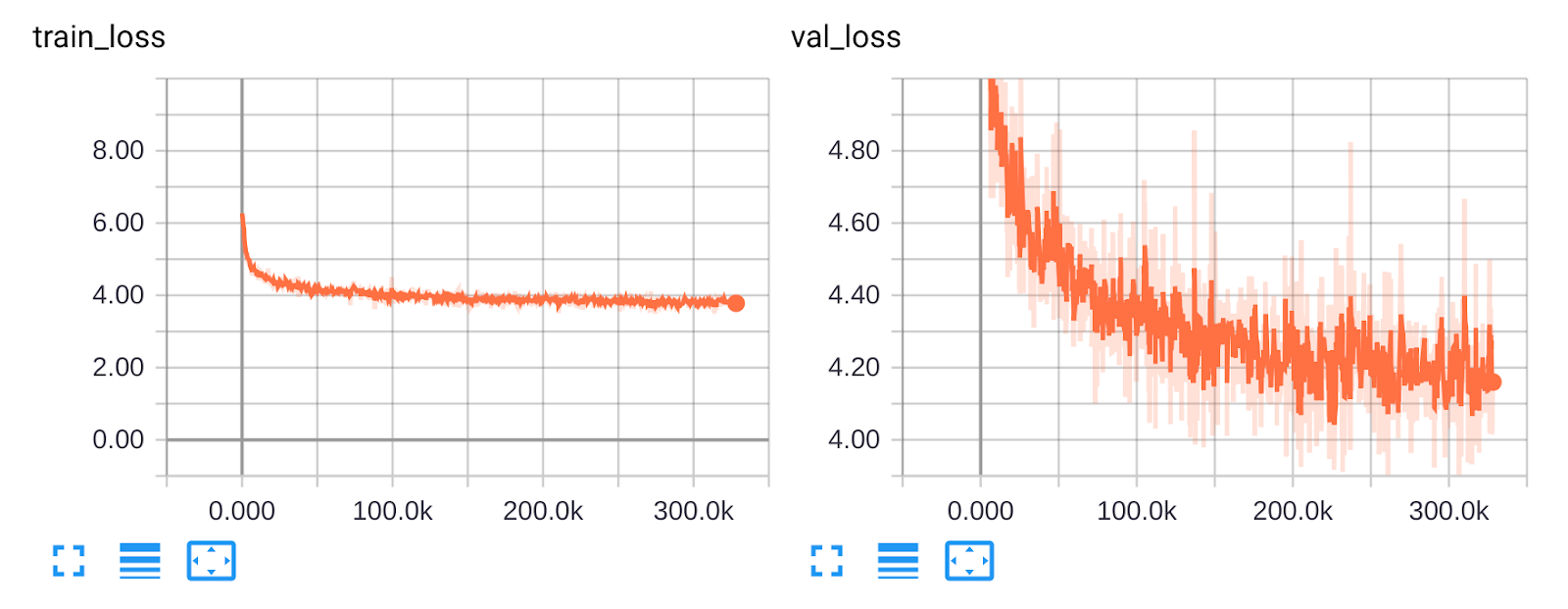}
\end{center}
\caption{Example loss values for both training and validation datasets, over the course of model training. Displayed for NTM-LM model specifically, but similar loss curves were observed for all models.}
\label{fig:training}
\end{figure*}

\section{Dual-NTM Seq2Seq Dialogue Architecture}

As a preliminary approach, we implement a dialogue generation system with segment-level memory manipulation. Segment-level memory refers to memory of sub-sentence level, which often corresponds to entity mentions, event mentions, and proper names, etc. We use NTM as the memory module, because it is more or less a default choice before specialized mechanisms are developed. Details of NTMs can be found in \newcite{DBLP:journals/corr/GravesWD14}.

As in the baseline model, the encoder and decoder each has an Gated Recurrent Unit (GRU) inside. A GRU is a type of recurrent neural networks that coordinates forgetting and write of information, to make sure they don't both occur simultaneously. This is accomplished via an "update gate." A GRU architecture processes a list of inputs in sequence, and is described by the following equations:

\begin{equation}
\begin{aligned}
z_t=\sigma (W_z \cdot [h_{t-1}, x_t]) \\
r_t=\sigma (W_r \cdot [h_{t-1}, x_t]) \\
\tilde{h_t}=\tanh(W \cdot [r_t * h_{t-1}, x_t]) \\
h_t=(1-z_t)*h_{t-1}+z_t*\tilde{h_t}
\end{aligned}
\end{equation}

\noindent
For each input $x_t$ and previous state $h_{t-1}$, the GRU produces the next state $h_t$ given learned weights $W_z$, $W_r$ and $W$. $z_t$ denotes the update gate. The encoder GRU in this memory architecture reads a token at each time step, and encodes a context representation $c$ at the end of the input sequence. In addition to that, the memory enhanced model implements two Neural Turing Machines (NTMs). Each of them is for one speaker in the conversation, since the Ubuntu dataset has two speakers in every conversation. 
Every turn in a dialogue is divided in 4 ``segments". If a turn has 20 tokens, for example, a segment contains 5 tokens. The output of the GRU is written to the NTM at the end of every segment. It does not output anything useful here, but the internal memory is being updated each time. When the dialogue switches to next turn, the current NTM pauses and the other NTM starts to work in the same way. When an NTM pauses, its internal memory retains, so as soon as the dialogue moves to its turn again, it continues to read and update its internal memory.
\begin{equation}\label{NTM_update}
    \mathrm{NTM}\leftarrow \mathrm{NTM}(s_{n\times T/4})
\end{equation}
Equation~\ref{NTM_update} shows how one NTM updates. $T$ denotes the length of one turn, and $s$ is the output of the encoder GRU. $n=1,2,3,4$ represents the four time steps when the NTM updates in one turn of the conversation.

The two NTMs can be interpreted as two external memories tracking each speaker's utterances. When one speaker needs to make a response at the end of the conversation, he needs to refer to both speakers' history to make sure the response is coherent with respect to context. This allows for separate tracking of each participant, while also consolidating their representations.

The decoder GRU works the same way as the baseline model. Each time it reads a token, from either the true response or the generated response, depending on whether teacher force training is used. This token and the context representation $c$ generated by the encoder GRU are both used as input to the decoder GRU. 

\begin{equation}
    s_t=\textrm{GRU}_{dec}([y_{t-1}\oplus c])
\end{equation}

However, now the two NTMs also participate in token generation. At every time step, output of the decoder GRU is fed into the two NTMs, and outputs of the two NTMs are used together to make predictions.

\begin{equation}
    \widehat{y_t}=\mathrm{softmax}(\mathrm{FC}(\mathrm{NTM_a}(s_{t-1})\oplus \mathrm{NTM_b}(s_{t-1})))
    \label{NTM_decoder}
\end{equation}

In the equation above, $\mathrm{FC}$ represents a fully connected layer, and $\widehat{y_t}$ is the predicted vector.

From now on, we refer to this system as the D-NTMS (Dual-NTM Seq2Seq) system.

\section{NTM Language Model Dialogue Architecture}

In this section we introduce a somewhat simpler, but more effective memory module architecture. In contrast to the previous D-NTMS architecture, we combine the encoder-decoder architecture of the sequence to sequence GRU into a single language model. This combination entails the model predicting all tokens in the dialogue history in sequence. This change in setup exploits the property that the response is in essence drawn from the same distribution as all previous utterances, and so should not be treated any differently. This language model variant learns to predict all utterances in the dialogue history, and thus treats the response as just another utterance to predict. This setup may also help the model learn the flow of conversation from beginning to end.

With a neural language model predicting tokens, it is then necessary to insert reads and writes from a Neural Turing Machine. In this architecture, we only use one NTM. This change is motivated by the possibility that the speaker NTMs from the previous architecture may have difficulty exchanging information, and thus cannot adequately represent each utterance in the context of the previous one. We follow an identical setup as before and split the dialogue history into segments. A GRU processes each segment in sequence. Between each segment, the output GRU state is used to query and write to the NTM module to store and retrieve relevant information about the context history so far. This information is conditioned on for all subsequent tokens in the next segment, in order to exploit this information to make more informed predictions. Lastly, the GRU NTM has an internal LSTM controller which guides the read and writes to and from the memory section. Reads are facilitated via content-based addressing, where a cosine similarity mechanism selects entries that most resemble the query. The Neural Turing Machine utilized can be found as an existing Github implementation\footnote{Impelementation for Neural Turing Machine can be found on Github here: https://github.com/loudinthecloud/pytorch-ntm}.

In further investigations, we refer to this model as the NTM-LM system.

\section{Baselines}

As a reliable baseline, we will evaluate a vanilla sequence-to-sequence GRU dialogue architecture, with the same hyper-parameters as our chosen model. We refer this this baseline as Seq2Seq. In addition, we report results for a vanilla GRU language model (LM). Finally, we include a more recent baseline, the Hierarchical Encoder-Decoder (HRED) system which is trained for the same number of epochs, same batch size, and with the same encoder and decoder size as the Seq2Seq baseline \footnote{We used the GRU-based HRED implementation available at \href{https://github.com/ctr4si/A-Hierarchical-Latent-Structure-for-Variational-Conversation-Modeling}{https://github.com/ctr4si/A-Hierarchical-Latent-Structure-for-Variational-Conversation-Modeling}}. As previously mentioned, we refer to our first proposed memory architecture as D-NTMS and to our second memory architecture as NTM-LM. 

\section{Evaluation}

To evaluate the performance of each dialogue baseline against the proposed models, we use the Ubuntu Dialogue Corpus \cite{lowe2015ubuntu}, chosen for its rich vocabulary size, diversity of responses, and dependence of each utterance on previous ones (coherence required). We perform perplexity evaluation using a held-out validation set. The results are reported in Table \ref{table:eval}. Perplexity is reported per word. For reference, a randomly-initialized model would receive a perplexity of 50,000 for our chosen vocabulary size. We also report generated examples from the model, shown in Table \ref{table:examples}.

\section{Results}

See Table \ref{table:eval} for details on model and baseline perplexity. To begin, it is worth noting that all of the above architectures were trained in a similar environment, with the exception of HRED, which was trained using an existing Github implementation implementation\footnote{Github imlementation of the HRED architecture can be found here: https://github.com/ctr4si/A-Hierarchical-Latent-Structure-for-Variational-Conversation-Modeling}. Overall, the NTM-LM architecture performed the best of all model architectures, whereas the sequence-to-sequence architecture performed the worst. The proposed NTM-LM outperformed the DNTM-S architecture.

After one epoch of training, the perplexity evaluated on the validation set was \emph{68.50} for the proposed memory-augmented NTM-LM architecture. This is a 0.68 perplexity improvement over the vanilla language model without the NTM augmentation.

\section{Discussion}

Overall, the HRED baseline was top performing among all tested architectures. This baseline breaks up utterances in a conversation and reads them separately, producing a hierarchical view which likely promotes coherence at a high level. 

Now we will discuss the memory-augmented D-NTMS architecture. The memory-augmented architecture improved performance above the baseline sequence-to-sequence architecture. As such, it is likely that the memory modules were able to store valuable information about the conversation, and were able to draw on that information during the decoder phase. One drawback of the memory enhanced model is that training was significantly slower. For this reason, model simplification is required in the future to make it more practical. In addition, the NTM has a lot of parameters and some of them may be redundant or damaging. In the DNTM-S system, we may not need to access the NTM at each step of decoding either. Instead, it can be accessed in some intervals of time steps, and the output is used for all steps within the interval.

The best performing model was the NTM-LM architecture. While the model received the best performance in perplexity, it demonstrated only a one-point improvement over the existing language model architecture. While in state-of-the-art comparisons a one point difference can be significant, it does indicate that the proposed NTM addition to the language model only contributed a small improvement. It is possible that the additional NTM module was too difficult to train, or that the NTM module injected noise into the input of the GRU such that training became difficult. It is still surprising that the NTM was not put to better use, for performance gains. It is possible the model has not been appropriately tuned.

Another consideration of the NTM-LM architecture is that it takes a significant amount of time to train. Similar to the D-NTMS, the NTM memory module requires a sizeable amount of computational steps to both retrieve a query response from available memory slots, and also to write to a new or existing slot using existing write weights. This must be repeated for each segment. Another source of slowdown with regard to computation is the fact that the intermittent NTM reads and writes force the input utterance into segments, as illustrated in Figure \ref{fig:model}. This splitting of token processing steps requires additional overhead to maintain, and it may discourage parallel computation of different GRU input segments simultaneously. This problem is not theoretical, and may be solved using future optimizations of a chosen deep learning framework. For Pytorch, we observed a slowdown for a segmented dialogue history versus a complete history.

Of all models, the HRED architecture utilized pre-trained GloVe vectors as an initialization for its input word embedding matrix. This feature likely improved performance of the HRED in comparison to other systems, such as the vanilla sequence-to-sequence. However, in separate experiments, GloVe vectors only managed a 5\% coverage of all words in the vocabulary. This low number is likely due to the fact that the Ubuntu Dialogues corpus contains heavy terminology from the Ubuntu operating system and user packages. In addition, the Ubuntu conversations contain a significant amount of typos and grammar errors, further complicating analysis. Context-dependent embeddings such as ElMo \cite{peters2018deep} may help alleviate this issue, as character-level RNNs can better deal with typos and detect sub word-level elements such morphemes.

Due to time requirements, there were no targeted evaluations of memory coherence other than perplexity, which evaluates overall coherence of the conversation. This form of specific evaluation may be achievable through a synethetic dataset of responses, for example, "What is your profession? I am a doctor.\textless /s\textgreater What do you do for work?\textless /s\textgreater I am a doctor." This sort of example would require direct storage of the profession of a given speaker. However, the Ubuntu Dialogue corpus contains complicated utterances in a specific domain, and thus does not lend well to synthesized utterances from a simpler conversational domain. In addition, synthetic conversations like the one above do not sound overly natural, as a human speaker does not normally repeat a query for information after they have already asked for it. In that sense, it is difficult to directly evaluate dialogue coherence.

Not reported in this paper was a separate implementation of the language model that achieved better results (62 perplexity). While this was the best performing model, it was written in a different environment than the language model reported here or the NTM-LM model. As such, comparing the NTM-LM to this value would be misleading. Since the NTM-LM is an augmentation of the existing LM language model implementation, we report perplexity results from that implementation instead for fair comparison. In that implementation, the addition of the NTM memory model improved performance. For completeness, we report the existence of the outperforming language model here.

\section{Conclusion}

We establish memory modules as a valid means of storing relevant information for dialogue coherence, and show improved performance when compared to the sequence-to-sequence baseline and vanilla language model. We establish that augmenting these baseline architectures with NTM memory modules can provide a moderate bump in performance, at the cost of slower training speeds. The memory-augmented architectures described above should be modified for increased computational speed and a reduced number of parameters, in order to make each memory architecture more feasible to incorporate into future dialogue designs.

In future work, the memory module could be applied to other domains such as summary generation. While memory modules are able to capture neural vectors of information, they may not easily capture specific words for later use. A possible future approach might combine memory module architectures with pointer softmax networks \cite{gulcehre2016pointing} to allow memory models to store information about which words from previous utterances of the conversation to use in future responses.

\bibliography{naaclhlt2019}
\bibliographystyle{acl_natbib}

\clearpage
\appendix

\section{Appendix}

\subsection{Preprocessing}

We construct a vocabulary of size 50,000 (pruning less frequent tokens) from the chosen Ubuntu Dialogues Corpus, and represent all missing tokens using a special unknown symbol \textless unk\textgreater. When processing conversations for input into a sequence-to-sequence based model, we split each conversation history into history and response, where response is the final utterance. To clarify, all utterances in each conversation history are separated by a special \textless \\s\textgreater symbol. A maximum of 170 tokens are allocated for the input history and 30 tokens are allocated for the maximum output response.

When inputting conversation dialogues into a language model-based implementation, the entire conversation history is kept intact, and is formatted for a maximum conversation length of 200 tokens. As for all maximum lengths specified here, an utterance which exceeds the maximum length is pruned, and extra tokens are not included in the perplexity calculation. This is likely not an issue, as perplexity calculations are per-word and include the end of sequence token.

\subsection{Training/Parameters}

All models were trained using the Adam optimizer \cite{kingma2014adam} and updated using a learning rate of 0.0001. All models used a batch size of 32, acceptable for the computational resources available. We develop all models within the deep learning framework Pytorch. To keep computation feasible, we train all models for one epoch. For reference, the NTM-LM architecture took over two and a half days of training for one epoch with the parameters specified.

\subsection{Layer Dimensions}

In our preliminary experiment, each of the NTMs in the D-NTMS architecture were chosen to have 1 read head and 1 write head. The number of memory slots is 20. The capacity of each slot is 512, the same as the decoder GRU state dimensionality. Each has an LSTM controller, and the size is chosen to be 512 as well. These parameters are consistent for the NTM-LM architecture as well.

All sequence-to-sequence models utilized a GRU encoder size of 200, with a decoder GRU size of 400. All language models used a decoder of size 400. The encoder hidden size of the HRED model was set to 400 hidden units. The input embedding size to all models is 200, with only the HRED architecture randomly initializing these embeddings with pre-trained GloVe vectors. The sequence-to-sequence architecture learns separate input word embeddings for encoder and decoder.

Each Neural Turing Machine uses 8 heads for reading and writing, with each head having a size of 64 hidden units In the case of the NTM-LM architecture, 32 memory slots are available for storage by the model. When breaking GPU computation to read and write from the NTM, we break the input conversation into segments of size 20 with NTM communication in-between segments. In contrast, the D-NTMS architecture uses a segment size of 5, and breaks up the conversation in utterances which fit in each segment.

\begin{table*}[ht]
\centering
\small
\begin{tabularx}{\linewidth}{ |X|X| } 
 \hline
Input Message History & Model Response \\
 \hline
\texttt{if the partition is below ubuntu , grub will complain as well as fstab .. but <unk> prefers to be \# ! partition </s> so i might get an error on boot after created a ntfs partion out of free space from ext3 <unk> and install xp on ( got ubuntu and grub installed ) . so neither os will boot and get error ? <eos>} & \texttt{eh.. i am running ubuntu off an texlive-base-bin and then i installed ubuntu after grub install 6.06 ( 11.10 ) is way useless i can ask again its nano broken} \\
\hline
\texttt{<unk> \# grub </s> i know what grub is but what is grub irc ? <eos>} & \texttt{italy : ) <eos> }\\
\hline
\texttt{hi , anyone can tell me how cani login as a root </s> just with a terminal , or in gnome ?} & \texttt{just that it does not appear <eos>} \\
\hline
\texttt{you can write a script to detect changes of your ip , and do action accordingly </s> do you have some how to ? i havent done bash scripting in years <eos>} & \texttt{how is the shell of the mean and when it could use ? wo while adding a while to reach them ? <eos>} \\
\hline
\texttt{have you tried another browser ? </s> : ya opera same problem <eos>} & \texttt{anyone else is recommend a lighter weight ubuntu release ? <eos>} \\
\hline
\texttt{when i check the error console there are many warnings </s> what browser are you using ? firefox ? <eos> } & \texttt{with firefox , can i paste my symptoms ? <eos>} \\
\hline
\texttt{nothing </s> is this on a default installation , or a long-time running system ? <eos>} & \texttt{what does that mean ? <eos>} \\
\hline
\texttt{virtualbox supports accelerated opengl in their latest version </s> does that mean i can play fifa sir ? </s> i have no idea .. you ' ll have to try .. it ' s still pretty new so there might be some bug
s <eos>} & \texttt{sure it is a closed bit more : p <eos>} \\
\hline
\texttt{it says : total <unk> . when i typed ls -ld <unk> . <unk> it says : total <unk> . drwx -- -- -- <unk> lisa lisa <unk> <unk> <unk> <unk> . <unk> . now ? </s> well , i would n't worry too much about this file right now , it looks ok to me . well , what were you ultimately trying to do ? <eos>} & \texttt{gui , apache2 . ufw controls that line in my terminal : sudo aptitude log output did n't <unk> close my user <eos>} \\
\hline
\texttt{how do i resolve my eth0 error : `` the interface does not exist `` </s> sudo ifconfig eth0 up </s> i ' m using eth0 now to talk . but system says it does n't exits . ( go figure ) <eos>} & \texttt{wifi0 is never a point to my knowledge <eos>} \\
\hline
\texttt{what ? find ~ ! -user \$ user </s> ... is a command <eos>} & \texttt{i looked for the pwd <eos>} \\
\hline

\end{tabularx}
\caption{Samples decoded with random sampling from the best-performing NTM-LM architecture. First column shows the message history, with the second column showing a model response. Due to the nature of the Ubuntu Dialogue Corpus, the terminology is complex.}
\label{table:examples}
\end{table*}

\end{document}